\documentclass{article}

\usepackage[hyphens]{url}
\usepackage[hidelinks]{hyperref}
\hypersetup{breaklinks=true}
\usepackage{amsmath}
\usepackage{times}
\usepackage{graphicx}
\usepackage{subfigure}
\usepackage{natbib}
\usepackage{algorithm}
\usepackage{algorithmic}
\usepackage{lipsum}
\usepackage{todonotes}
\usepackage[accepted]{icml2017}
\usepackage{float}

\icmltitlerunning{Conditional Variational Autoencoder for Neural Machine Translation}
\newcommand{\V}[1]{\mathbf{#1}}

\begin{document}

\twocolumn[
\icmltitle{Conditional Variational Autoencoder for Neural Machine Translation}
\begin{icmlauthorlist}
    \icmlauthor{Artidoro Pagnoni$^*$}{Harvard}
    \icmlauthor{Kevin Liu$^*$}{Harvard}
    \icmlauthor{Shangyan Li$^*$}{Harvard}
\end{icmlauthorlist}

\icmlaffiliation{Harvard}{School of Engineering and Applied Sciences, Harvard University}
\icmlcorrespondingauthor{Artidoro Pagnoni}{apagnoni@college.harvard.edu}
\vskip 0.3in
]
\printAffiliationsAndNotice{\icmlEqualContribution} 

\begin{abstract}
   We explore the performance of latent variable models for conditional text generation in the context of neural machine translation (NMT).
   Similar to \cite{vnmt}, we augment the encoder-decoder NMT paradigm by introducing a continuous latent variable to model features of the translation process. We extend this model with a co-attention mechanism motivated by \cite{parikh} in the inference network. Compared to the vision domain, latent variable models for text face additional challenges due to the discrete nature of language, namely posterior collapse \cite{bowman}. We experiment with different approaches to mitigate this issue.
   We show that our conditional variational model improves upon both discriminative attention-based translation and the variational baseline presented in \cite{vnmt}. Finally, we present some exploration of the learned latent space to illustrate what the latent variable is capable of capturing. This is the first reported conditional variational model for text that meaningfully utilizes the latent variable without weakening the translation model.

\end{abstract}

\section{Introduction}
\label{sec:introduction}




Machine translation is a classic, conditional language modeling task in NLP, and was one of the first in which deep learning techniques trained end-to-end have been shown to outperform classical phrase-based pipelines. Current NMT models generally use the encoder-decoder framework \cite{sutskever} where an encoder transforms a source sequence to a distributed representation, which the decoder then uses to generate the target sequence. Additionally, attention mechanisms \cite{bahdanau} allow the model to focus on relevant parts of the source sequence when decoding. However, these attention-based models may be insufficient in capturing all alignment and source sentence information \cite{tu}.

To attempt to more fully capture holistic semantic information in the translation process, we explore latent variable models. Latent variable models are a class of statistical models that seek to model the relationship of observed variables with a set of unobserved, latent variables, and can allow for modeling of more complex, generative processes. However, inference in these models can often be difficult or intractable, motivating a class of variational methods that frame the inference problem as optimization. Variational Autoencoders \cite{kingma}, in particular, have seen success in tasks such as image generation \cite{draw}, but face additional challenges when applied to discrete tasks such as text generation \cite{bowman}.

We experiment with a conditional latent variable model applied to the task of translation. \cite{vnmt} introduce a framework and baseline for conditional variational models and apply it to machine translation. We extend their model with a co-attention mechanism, motivated by \cite{parikh}, in the inference network and show that this change leads to a more expressive approximate posterior. We compare our conditional variational model with a discriminitive, attention-based baseline, and show an improvement in BLEU on German-to-English translation. We also present our experiments testing various methods of addressing common challenges of applying VAEs to text \cite{bowman}, namely posterior collapse. Finally, we demonstrate some exploration of the learned latent space in our conditional variational model.

\section{Background}
This section discusses recent efforts in neural machine translation, variational autoencoders (VAE), and their extension to the conditional case (CVAE).

\subsection{RNN-Attention Sentence Encoding}
In the standard Recurrent Neural Net (RNN)-based encoder-decoder setting, the encoder RNN represents the source sentence by learning sequentially from the previous source word $x_i$ and an evolving hidden state, while the decoder RNN similarly predicts the next target word $y_i$ using the previously generated output and its own hidden state. The probabilistic decoder model seeks to maximize $p(\mathbf{y}|\mathbf{x})$, the likelihood of output sequence $\mathbf{y}$ given source input $\mathbf{x}$. The attention mechanism introduced in \cite{bahdanau} enhances this model by aligning source and target words using the encoder RNN hidden states. However, it has been shown that this type of models struggles to learn smooth, interpretable global semantic features \cite{bowman}.
\subsection{Variational Autoencoder}
The variational autoencoder (VAE) \cite{kingma} is a generative model that uses deep neural nets to predict parameters of the variational distribution. This models the generation of $\mathbf{y}$ as conditioned on an unobserved, latent variable $\mathbf{z}$ by $p_{\theta}(\mathbf{y}|\mathbf{z})$ (where $\theta$ represents parameters in the neural network), and seeks to maximize the data log likelihood $p_{\theta}(\V{y})$. The main principle of VAE is to introduce an approximate posterior $q_{\phi}(\V{z}|\V{y})$ with variational parameters predicted by a neural network, in order to address the intractability of the true posterior $p_\theta({\V{z|y}})$ in maximum likelihood inference. It can be seen as a regularized version of an autoencoder, where $q_\phi(\V{z|y})$ can be considered the encoder and $p_\theta(\V{y|z})$ the decoder. The objective is: 

\begin{align*}
    \log p_{\theta}(\V{y})\geq&\int \log p(\V{y}|\V{z})p(\V{z})d\V{z}\\
    =&E_{q_{\phi}(\V{z}|\V{y})}\left[\log\frac{p(\V{y}|\V{z})p(\V{z})}{q(\V{z|y})}\right]\\
    =&-KL(q_{\phi}(\V{z|y})|p_{\theta}(\V{z}))+E_{q_{\phi}(\V{z}|\V{y})}[\log p_{\theta}(\V{y|z})]\\
    \equiv&\mathcal{L}_{\textnormal{VAE}}(\theta,\phi;\V{y})
\end{align*}

Gradients for this objective, also called the Evidence Lower Bound (ELBO), can be estimated with Monte Carlo sampling and the reparameterization trick \cite{kingma, rezende}. Training can then be done end-to-end with stochastic gradient optimization. To generate new samples, a latent variable $\mathbf{z}$ can be drawn from the prior $p(\mathbf{z})$, then the sample can be generated from $p_{\theta}(\mathbf{y}|\mathbf{z})$.

\subsection{Conditional Variational Autoencoder}
Conditional variational autoencoder (CVAE) is an extension of VAE to conditional tasks such as translation. Each component of the model is conditioned on some observed $\mathbf{x}$, and models the generation process according to the graphical model shown below.

\begin{center}
\includegraphics[]{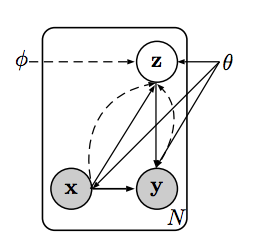}

Graphical Model of CVAE

Solid lines denote the generation process and dashed lines denote the variational approximation.

Figure from \cite{vnmt}
\end{center}

CVAE seeks to maximize $\log p_\theta(\V{y|x})$, and the variational objective becomes:

\begin{align*}
    \log p_\theta(\V{y|x})\geq&\int\log p(\V{y|x,z})p(\V{z|x})d\V{z}\\
    =&-{KL}(q_\phi(\V{z|x,y})|p_\theta(\V{z|x}))\\
    &+{E}_{q_\phi}[\log p(\V{y|x,z})]\\
    \equiv&\mathcal{L}_{\textnormal{CVAE}}(\theta,\phi;\V{x,y})
\end{align*}

Here, CVAE can be used to guide NMT by capturing features of the translation process into the latent variable $\mathbf{z}$.

\section{Related Work} 
There has been substantial exploration on both the neural machine translation and variational autoencoder fronts. The attention mechanism introduced by \cite{bahdanau} has been extensively used with RNN encoder-decoder models \cite{DBLP:journals/corr/WangJ15b} to enhance their ability to deal with long source inputs. 

\cite{bowman} presents a basic RNN-based VAE generative model to explicitly model holistic properties of sentences. It analyzes challenges for training variational models for text (primarily posterior collapse) and propose two workarounds: 1. KL cost annealing and 2. masking parts of the source and target tokens with $\texttt{'<unk>'}$ symbols in order to strengthen the inferer by weakening the decoder ("word dropouts"). This model is primarily concerned with unconditional text generation and does not discuss conditional tasks.

\cite{savae} is one of the first LSTM generative models to outperform language models by using a latent code. It proposes a hybrid approach between amortized variational inference (AVI) to initialize variational parameters and stochastic variational inference (SVI) to iteratively refine them . The proposed approach outperforms strong autoregressive and variational baselines on text and image datasets, and reports success in preventing the posterior-collapse phenomenon.

\cite{vnmt} introduces the basic setup for a conditional variational language model and applies it to the task of machine translation. It reports improvements over vanilla neural machine translation baselines on Chinese-English and English-German tasks.\\


\section{Model}


The model that we propose relies on an encoder-decoder translation architecture similar to \cite{bahdanau} along with an inferer network. In order to reduce the number of parameters to be trained as well as to avoid overfitting, we share embeddings and RNN parameters between the translation and the inferer networks.

\subsection{Neural Encoder}
We use a bidirectional LSTM \cite{DBLP:journals/neco/HochreiterS97} to produce annotation vectors for words in both the source sentence $x$ and the target sentence $y$. The LSTM outputs from the forward and backward passes are concatenated to produce a unique annotation vector for each word.
\begin{eqnarray*}
 &\V{h^x_t} =(\mathrm{LSTM}(E_x(x_{t}), \overrightarrow{\V{h}}^x_{t-1} ); \mathrm{LSTM}(E_x(x_{t}), \overleftarrow{\V{h}}^x_{t+1} ))\\  
 &\V{h^y_t} = (\mathrm{LSTM}(E_y(y_{t}), \overrightarrow{\V{h}}^y_{t-1} ); \mathrm{LSTM}(E_y(y_{t}), \overleftarrow{\V{h}}^y_{t+1} ))
\end{eqnarray*}

Where $E_x$, and $E_y$ are learned source and target word embeddings. Although word embeddings are already continuous representations of words, the additional LSTM step introduces contextual information that is unique to the sentence.   

\subsection{Neural Inferer}
The neural inferer can be divided in two parts: the prior and the posterior networks. Both prior and posterior distributions are assumed to be multivariate Gaussians. As determined by the ELBO equation, the parameters of the prior are computed by the prior network which only takes the source sentence as input. The posterior parameters are determined from both the source and the target sentences. We restrict the variance matrices of the prior and the posterior distributions to be diagonal.

\subsubsection{Neural Prior}
The prior distribution, denoted 
$$p_\theta(\V{z|x}) = \mathcal{N}(\V{z}; \mu(\V{x}), \Sigma^2(\V{x}))$$
is a multivariate Gaussian with mean and variance matrices parametrized by neural networks. We use the same network architecture proposed in VNMT \cite{vnmt}.

The source, a variable length sentence, is mapped to two fixed dimensional vectors, the mean and the variance of the multivariate gaussian distribution. First, we obtain a fixed dimensional representation of the sentence by mean-pooling the annotation vectors produced by the neural encoder over the source sentence. Then we add a linear projection layer $W_z$, and a non linearity.
$$\V{h_z} = \tanh\left(W_z\left(\frac{1}{T_x} \sum_t \V{h_t}\right) + \V{b_z}\right)$$
We finally project to the mean vector and the scale vector:
\begin{eqnarray*}
\V{\mu} &=& W_\mu \V{h_z} + \V{b_\mu}\\
\log\Sigma^2 &=& I~(W_\sigma \V{h_z} + \V{b_\sigma})
\end{eqnarray*}
Where $I$ is the identity matrix. We explored concatenating a self-attention context vector to the mean-pool of the annotation vectors. This addition did not alter the performance of the model and we decided not to include it in the final model for which we report results bellow. Although \cite{parikh} proposed self-attention as a fixed-size representation of a sentence, our results indicate that mean-pooling the annotation vectors encodes similar information.

\subsubsection{Neural Posterior}
During training, the latent variable will be sampled from the posterior distribution:
$$q_\phi(\V{z|x,y}) = \mathcal{N}(\V{z}; \mu'(\V{x,y}), \Sigma^2(\V{x,y}))$$
a multivariate Gaussian, with parameters depending on both source and target sentences.

We introduce a new architecture for the neural posterior inspired by Parikh's co-attention  \cite{parikh}. In the context of variational autoencoders, it is crucial that the posterior network is as expressive as possible. We found that the posterior used in VNMT \cite{vnmt}, which simply takes the concatenated mean-pool vectors of the source and target codes, does not capture interactions between the source and the target sentences. Intuitively, having access to both sentences introduces the possibility of finding important stylistic translation decisions by comparing the two sentences. There are many ways in which a sentence can be translated due to the multimodal nature of natural language, and latent variable models aim at capturing precisely these translation specificities through the latent variable. Not capturing source-target interactions is thus a serious drawback in a CVAE model, we propose the first model with such interactions.

In the same spirit as the co-attention technique described in \cite{parikh}, we compute pairwise dot attention coefficients between the words of the source sentence and each word of the target sentence, and vice versa. Notice here that instead of applying the co-attention mechanism directly to the word embeddings as it is done in \cite{parikh}, we apply it to the annotation vectors produced by running the encoder LSTM on both source and target sentences. We found that this approach lead to a more representative posterior network, which gave better results. 

The source and target attention coefficients are therefore given by:
\begin{eqnarray*}
&\V{\alpha}^x_t = \frac{\exp(\V{h}^y_t\V{h}^{x\top}_i)}{\sum_i \exp(\V{h}^y_t\V{h}^{x\top}_i)} = \mathrm{softmax}(\V{h}^y_t\V{h}^{x\top})\\
&\V{\alpha}^y_t = \frac{\exp(\V{h}^x_t\V{h}^{y\top}_i)}{\sum_i \exp(\V{h}^x_t\V{h}^{y\top}_i)} = \mathrm{softmax}(\V{h}^x_t\V{h}^{y\top})
\end{eqnarray*}
Where the softmax is take over the second dimension. We then use these coefficients to get context vectors, which are convex combinations of the annotation vectors:
\begin{eqnarray*}
&\V{c}^x_t = \V{\alpha}^x_t \V{h}^{x}\\
&\V{c}^y_t = \V{\alpha}^y_t \V{h}^{y}
\end{eqnarray*}
We combine the previous with a mean-pool to obtain a fixed dimensional vector, and concatenate it with the mean-pool of the annotation vectors from both the source and target sentences (similar to what is done in the prior). 
\begin{eqnarray*}
&\bar{\V{c}^x} = \frac{1}{T_f} \sum_{t=1}^{T_f} \V{c}^x_t\\
&\bar{\V{c}^y} = \frac{1}{T_s} \sum_{t=1}^{T_s} \V{c}^y_t\\
&\bar{\V{h}^x} = \frac{1}{T_s} \sum_{t=1}^{T_s} \V{h}^x_t\\
&\bar{\V{h}^y} = \frac{1}{T_f} \sum_{t=1}^{T_f} \V{h}^y_t
\end{eqnarray*}
Finally, we add a linear projection layer and a non linearity, and get the final fixed dimensional vector.
$$\V{h}'_z = \tanh(W'_z (\bar{\V{c}^x};\bar{\V{h}^x};\bar{\V{c}^y};\bar{\V{h}^y}) + \V{b}'_z)$$
This will be projected to the mean vector and variance matrix just like in the prior network:
\begin{eqnarray*}
\V{\mu}' &=& W'_\mu \V{h}'_z + \V{b}'_\mu\\ 
\log\Sigma'^2 &=& I(W'_\sigma \V{h}'_z + \V{b}'_\sigma)\\
\end{eqnarray*}

Through the use of the co-attention network, the mean and variance parameters of the posterior capture interactions between source and target sentences.

\subsection{Neural Decoder}
The decoder models the probability of a target sentence $\V{y}$ given a source sentence $\V{x}$ and a latent variable $\V{z}$ by decomposing the generation process in a left to write process. At each time step given $\V{y}_{<j}$, the words that were already translated, $\V{x}$ and, $\V{z}$ the decoder outputs a probability distribution over the vocabulary.
$$p(\V{y|x,z}) = \prod_{j=1}^{T_f}p(\V{y}_{j}|\V{y}_{<j}, \V{x, z})$$
We use Bahdanau's attention decoder \cite{bahdanau} with the incorporation of the dependence on the latent variable $z$. In particular we can parametrize the probability of decoding each word as:
$$p(\V{y}_{j}|\V{y}_{<j}, \V{x, z}) = \mathrm{softmax}(W_v \tanh(\V{h}_j; \V{c}_j; \V{z}) + \V{b}_v)$$
Where $W_v$ is a linear projection to a vocabulary-sized vector, $\V{h}_j$ is the output of the LSTM at step $j$, $\V{c}_j$ is the context vector for time step $j$, and $\V{z}$ is the sentence level latent variable. 

The context vector $\V{c}_j$ is the result of a convex combination of the annotation vectors $\V{h}^x$ produced by the encoder applied to the source sentence $\V{x}$.
$$\V{c}_j = \V{\alpha}_j \V{h}^{x}$$
Where $\V{\alpha}_j$ is the vector of normalized attention weights obtained by taking the softmax of the dot product of annotation vectors and the LSTM output $\V{h}_j$. 
$$\V{\alpha}_j = \mathrm{softmax}(\V{h}_j \V{h}^{x\top})$$
The hidden state $\V{h}_j$ is produced at each step by a LSTM that takes as input $\V{z}$ and the word embedding of word $\V{y}_j$.
$$\V{h}_j = \mathrm{LSTM}((E_y(\V{y}_j);\V{z}), \V{h}_{j-1})$$

The decoder network that we present differs from Bahdanau's architecture in that we include the dependency on the latent variable $z$. The vector $z$ is concatenated before the last projection layer to the context vector and the LSTM hidden state. We also included it as a skip connection in the LSTM input by concatenating it to the word embedding of the target words at each time step.







\section{Methods}

We use the IWSLT 2016 German-English dataset for our experiments, consisting of 196k sentence pairs. We preprocess by filtering out pairs containing sentences longer than 100 words and replacing all words that appear less than five times with an "unk" token, yielding vocabulary sizes of 26924 German words and 20489 English words. Note: for BLEU score calculation in our current results, we retain the "unk" tokens and thus may not be directly comparable to other published results.

We trained each of our models end-to-end with Adam \cite{adam} with initial learning rate 0.002, decayed by a scheduler on plateau. Our variational models used Monte Carlo sampling and the reparameterization trick for gradient estimation \cite{kingma, rezende}. Latent variables are sampled from the approximate posterior during training, but from the prior during generation.

For our variational models, we use a KL warm-up schedule by training a modified objective: $$J = \text{RE} + \alpha \text{KL}$$ Alpha is set to $0$ for the first five training epochs, then annealed linearly over the next ten epochs.

We compared three models: vanilla sequence-to-sequence with dot-product attention, VNMT \cite{vnmt}, and our Conditional VAE with co-attention. All models used 300 dimensional word embeddings, 2 layer encoder and decoder LSTMs with hidden dimensions of size 300. Variational models used 32 dimensional latent variables.





\section{Results}

Our main results comparing discriminative attention-based translation with a few of our CVAE models are shown in Table1.

\begin{table*}[t!]
 \centering
 \begin{center}
   \begin{tabular}{|| c | c | c | c | c | c | c ||} 
   \hline
   Model & PPL & NELBO/NLL & RE & KL & BLEU Greedy & BLEU \\ [0.5ex] 
   \hline\hline
   Seq2seq & 7.7103 & 2.0426 & NA & NA & 28.43 & 30.22 \\ 
   \hline
   CVAE & $\leq$ \bf{7.6879} & 2.0397 & 2.0305 & 0.0092 & 29.94  & \bf{31.2} \\
   \hline
   CVAE - KL coeff = 0.25 & $\leq$ 9.275 & 2.2273 & \bf{1.7733} & \bf{0.4540} & 29.21 & \bf{30.96} \\
   \hline
  \end{tabular}
\end{center}
 \caption{Experiment Results: Main Results}
 Perplexity, Negative ELBO / Negative Log Likelihood, Reconstruction Error, KL per Word, and BLEU scores for generation with greedy search, greedy search with zeroed out latent variable, and beam search with width 10.
 \label{fig:mainres}
\end{table*}

\subsection{Experiment 1: Expressiveness of Co-attention Inference}
To assess the contribution of our co-attention based approximate posterior, we compare the reconstruction losses of our model and the VNMT model \cite{vnmt} with the KL term of the ELBO objective zeroed out. Here, all gradients will only be backpropagated through the reconstruction error, eliminating the KL regularization of the approximate posterior to resemble the prior. Hence, the reconstruction error here is a measure of the ability of the approximate posterior to encode information relevant to reconstructing the target sequence. Results are shown below in Table 2.
\begin{table}[H]
 \centering
 \begin{center}
   \begin{tabular}{|| c | c ||} 
   \hline
   Model & RE \\ [0.5ex] 
   \hline\hline
   VNMT &  1.5771 \\
   \hline
   Co-attention & 1.3572\\ 
   \hline
  \end{tabular}
\end{center}
 \caption{Experiment 1: Expressiveness of Co-attention Inference}
 \label{fig:exp1}
\end{table}

\subsection{Experiment 2: Addressing Posterior Collapse}

Next we explore three methods of addressing posterior collapse: Word Dropout, KL Minimum, and KL Coefficient. Results are shown in Table 3.

\subsubsection{Word Dropout}
Extending word dropout as used in \cite{bowman}, we weaken the encoder-decoder portion of the model to steer the model to make greater use of the latent variable when translating. We mask words with $\texttt{<unk>}$ in both the source and target sequences before feeding them into the encoder and decoder, respectively. However, we do not mask words fed into the inference networks, hoping to more strongly incentive use of the the latent variable.

\subsubsection{KL Minimum}
We set a minimum KL penalty in the objective, forcing the model to take at least a fixed KL regularization cost.
$$J = \textnormal{RE} + \text{max}(\textnormal{KL}, \textnormal{KL}_{\textnormal{MIN}})$$

\subsubsection{KL Coefficient}
We fix a constant coefficient $\V{\alpha}_{KL}$ to the KL objective, allowing us to adjust the weighting of the KL penalty relative to reconstruction error.
$$J = \textnormal{RE} + \V{\alpha}_{\textnormal{KL}} \textnormal{KL} $$

\begin{table*}[t!]
 \centering
 \begin{center}
   \begin{tabular}{|| c | c | c | c | c | c ||} 
   \hline
   Model & PPL & NELBO/NLL & RE & KL & BLEU Greedy\\ [0.5ex] 
   \hline\hline
   CVAE & $\leq$ 7.687 & 2.0397 & 2.0305 & 0.0092 & 29.94\\
   \hline
   CVAE - min KL = 0.1 & $\leq$ 7.741 & 2.0466 & 1.9677 & 0.0788 & 29.15\\
   \hline
   CVAE - min KL = 0.2 & $\leq$ 8.031 & 2.0833 & 1.9294 & 0.1539 & 28.85\\
   \hline
   CVAE - KL coeff = 0.1 & $\leq$ 14.323 & 2.6619 & 1.6203 & 1.0416 & 29.16\\
   \hline
   CVAE - KL coeff = 0.25 & $\leq$ 9.275 & 2.2273 & 1.7733 & 0.4540 & 29.21\\
   \hline
  \end{tabular}
\end{center}
 \caption{Experiment 2: Addressing Posterior Collapse}
  Perplexity, Negative ELBO / Negative Log Likelihood, Reconstruction Error, KL per Word, and BLEU scores for generation with greedy search, greedy search with zeroed out latent variable, and beam search with width 10.
 \label{fig:exp2}
\end{table*}

\subsection{Experiment 3: Generation and Interpolation}
To explore the latent space learned by the model, we sample and generate multiple sequences. To verify that the latent space is smooth, we interpolate across the latent space and observe sentences generated. Figure 1 shows 20 sampled sentences for each example, ranked by log probability. Figure 2 shows examples of linear interpolations between two sampled latent variables.
These experiments are done with the CVAE model trained with KL coefficient of 0.25.

\begin{figure*}[t!]
\begin{center}
\includegraphics[scale=0.6]{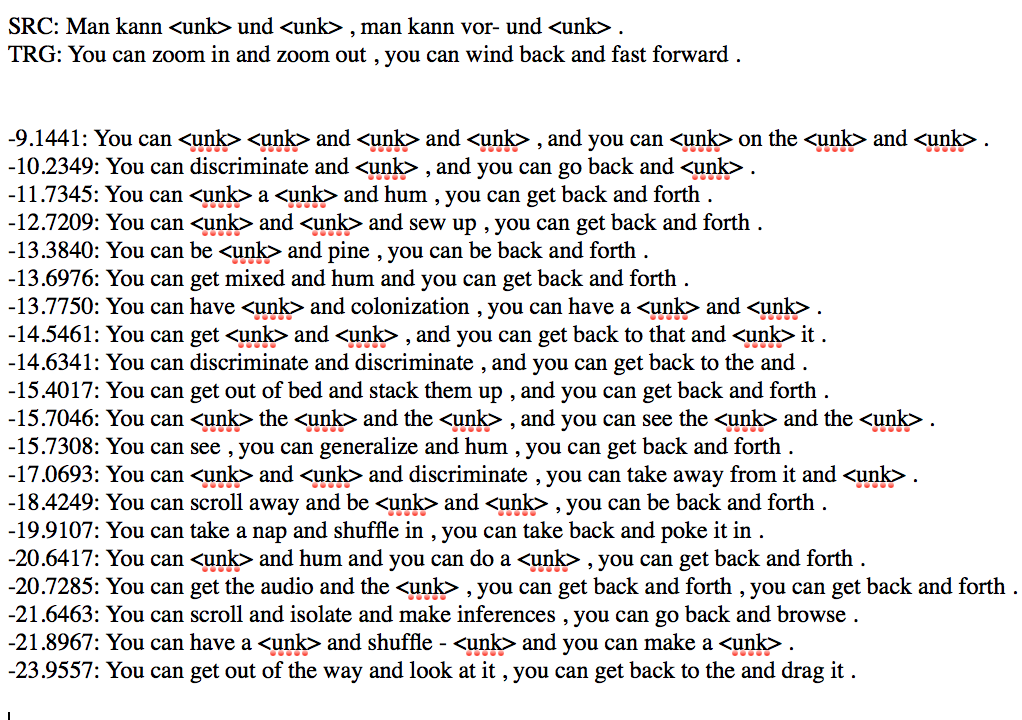}
\includegraphics[scale=0.6]{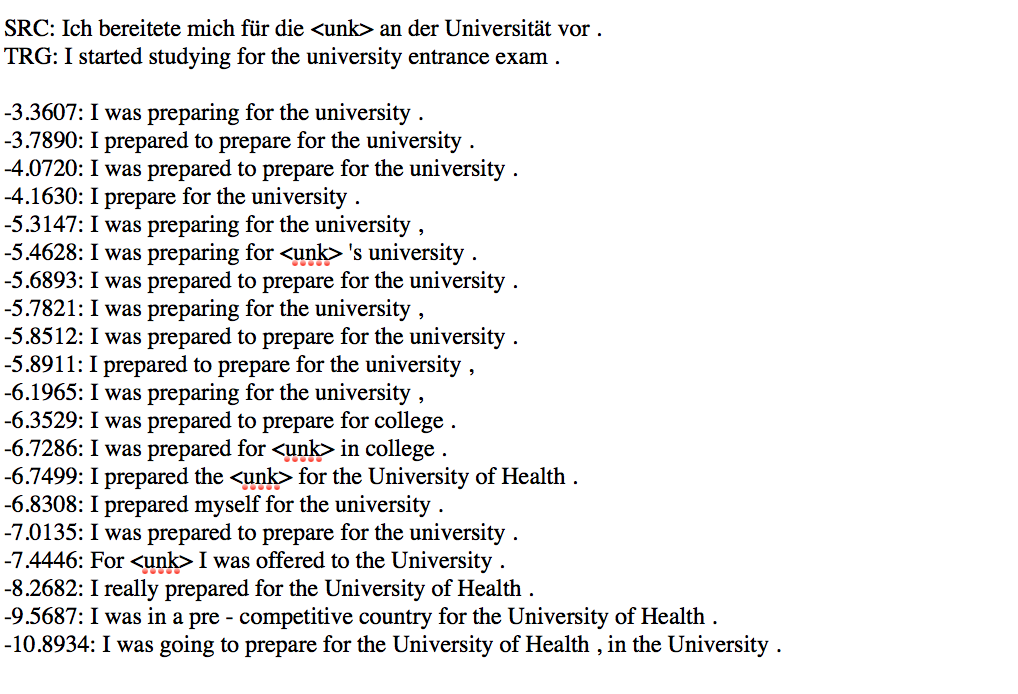}
\includegraphics[scale=0.6]{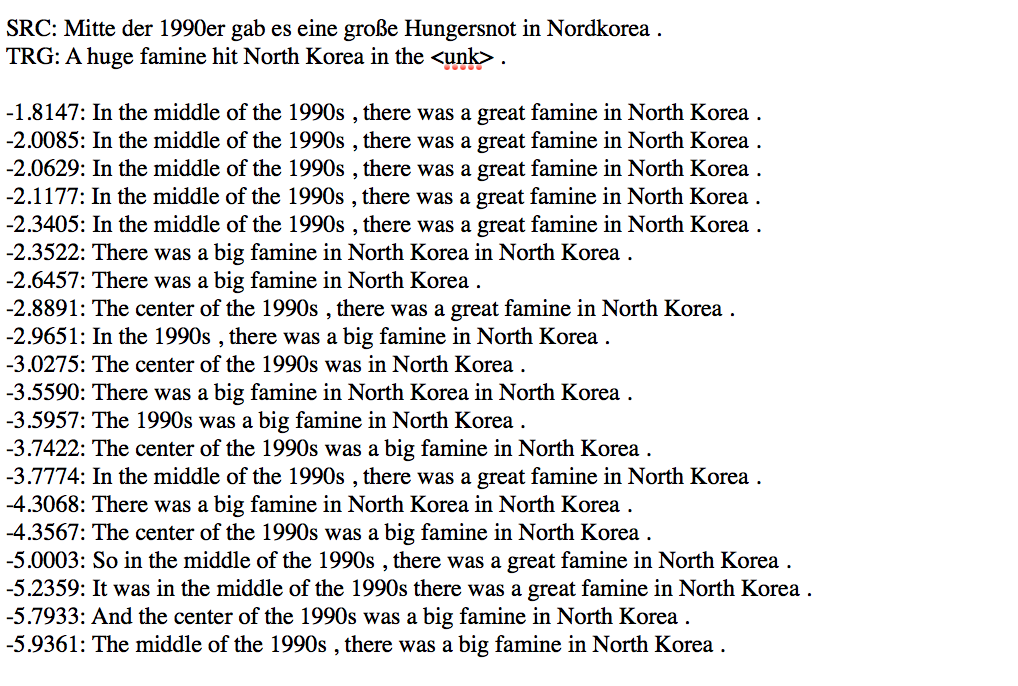}
\caption{Experiment 3: Ranked generation of samples by log probabilities}

Generated sentences from sampling from the prior. Ranked by log likelihood.
\end{center}
\end{figure*}

\begin{figure*}[t!]
\begin{center}
\includegraphics[scale=0.6]{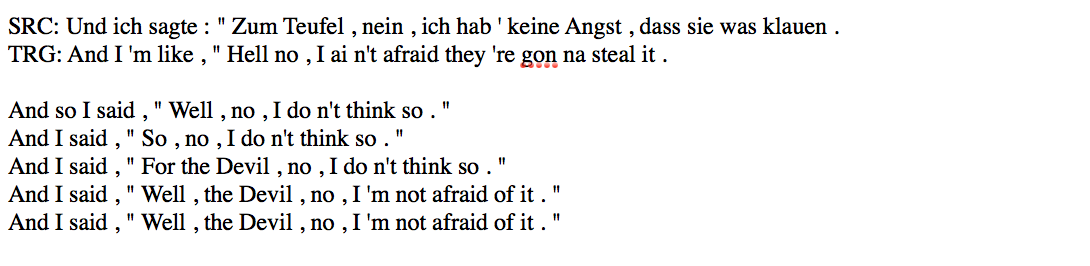}
\includegraphics[scale=0.6]{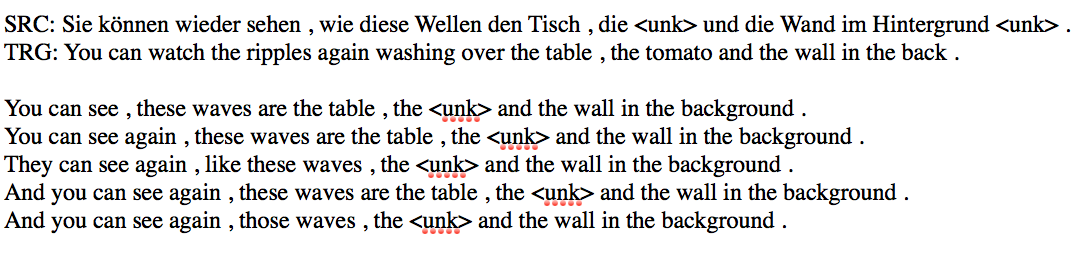}
\includegraphics[scale=0.6]{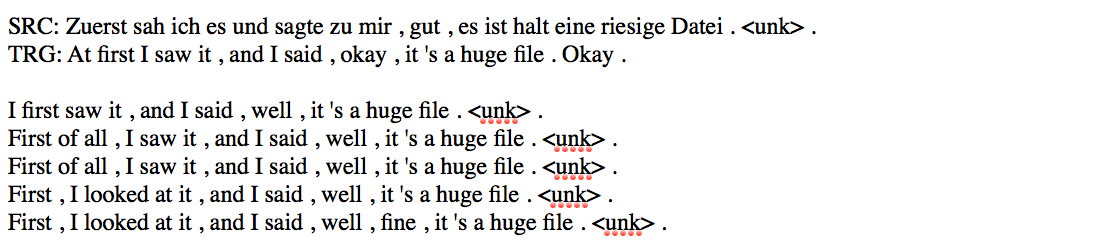}
\caption{Experiment 3: Interpolations}

Linear interpolations between two sampled latent variables.
\end{center}
\end{figure*}






\section{Discussion}

From our main results, our variational models are able to outperform a vanilla sequence-to-sequence model with attention by both BLEU and perplexity measures. However, as expected with VAEs for text, we ran into the challenge of posterior collapse for our standard CVAE model. By setting KL coefficient to 0.25 (described above), we are able to train a model that utilizes the latent variable model much more, and still outperform sequence-to-sequence in terms of BLEU.

In experiment 1, we show that the addition of our co-attention mechanism significantly improves the expressiveness of the approximate posterior network. This indicates the potential for our CVAE model to improve on previous variational baselines for translation. Furthermore, this result also confirms that capturing interactions between source and target sentences through co-attention helps provide effective information to the latent variable about the specificities of the translation process.   

In experiment 2, we present a comparison between various methods of combating posterior collapse. As expected, there is a trade-off between reconstruction error and KL. Although most recent work on VAE for text in the unconditional setting has focused on various methods of weakening the decoder to face posterior collapse, we show that modifying the learning objective incentivizes the use of the inference network without affecting the translation quality. We explored with setting a minimum for the KL, adding a coefficient to the KL penalty term in the ELBO, and word dropout. 

When setting a minimum for the KL, we essentially provide a minimum budget of KL that the inference network can use. In the unconditional setting, minimum KL budgeting can be achieved through the use of von Mises Fisher distribution, with uniform prior. However, in the conditional setting, with a prior that is not uniform this approach is not viable. The principal issue with setting an explicit minimum to the KL term is that when the KL term is smaller than the predefined value, there is no gradient propagated through the KL objective. The posterior is still updated through the reconstruction error term, but the prior is not updated, as it only appears in the KL term.

In experiment 3, we show an exploration of the latent space.

From sampling and generating (Figure 1), we observe that the model is able to produce somewhat diverse sentences. In the first example, the source sentence contains several \texttt{<unk>} tokens and thus there is a lot of uncertainty to what the sentence could mean. The generated samples are quite diverse, mentioning topics such as shuffling, beds, colonization, discrimination, etc. This demonstrates that latent variables could encode diverse semantic information. In the second example, there are slight variations in tense: "prepare", "prepared", "was preparing", etc. The third
example shows variation in wording: "In the middle of the 1990s", "center of the 1990s", "In the 1990s", etc. These examples illustrate some of the semantic and stylistic attributes of the translation process that can be captured by the latent variable.

From our interpolations (Figure 2), we see that the model is able to learn reasonably smooth latent representations for translations. 

From these explorations, we confirm that the model is learning a meaningful and smooth latent space that can guide the translation process.




\section{Conclusion}

We propose a conditional variational model for machine translation, extending the framework introduced by \cite{vnmt} with a co-attention based inference network and show improvements over discriminitive sequence-to-sequence translation and previous variational baselines. We also present and compare various ways of mitigating the problem of posterior collapse that has plagued latent variable models for text. Finally, we explore the latent space learned and show that it is able to represent somewhat diverse sentences and smooth interpolations. Future work includes further exploration of latent spaces for text, applying the model to larger translation datasets or other conditional tasks such as summarization, observing performance by sentence length, and closer analysis into what the latent variable is contributing.





\end{document}